\documentclass[11pt]{article}

\usepackage[preprint]{acl}

\usepackage{times}
\usepackage{latexsym}

\usepackage[T1]{fontenc}

\usepackage[utf8]{inputenc}

\usepackage{microtype}

\usepackage{inconsolata}

\usepackage{graphicx}
\usepackage{hyperref}       
\usepackage{url}            
\usepackage{booktabs}       
\usepackage{amsfonts}       
\usepackage{nicefrac}       
\usepackage{microtype}      
\usepackage{xcolor}         
\usepackage{amsmath}
\usepackage{siunitx}    
\usepackage{caption}
\usepackage{graphicx}
\usepackage{colortbl}
\usepackage{array}
\usepackage{wrapfig} 
\usepackage{colortbl} 
\usepackage{tcolorbox}
\title{OPERA: Aligning Open-Ended Reasoning via Objective Perplexity-based Reinforcement Learning}

\author{
\textbf{Wenxuan Jiang\textsuperscript{1,2}\thanks{~~Work done during an internship at Meituan.}},
 \textbf{Zining Fan\textsuperscript{3}},
 \textbf{Zijian Zhang\textsuperscript{2}\thanks{~~Corresponding author.}},
 \textbf{Xuecheng Wu\textsuperscript{4}},\\
 \textbf{Hongming Tan\textsuperscript{3}}, 
 \textbf{Haoyang Dai\textsuperscript{5}},
 \textbf{Xiaoyu Li\textsuperscript{2}},
 \textbf{Xuezhi Cao\textsuperscript{2}},
 \textbf{Ninghao Liu\textsuperscript{1}\footnotemark[2]},
\\
 \textsuperscript{1}The Hong Kong Polytechnic University
 \textsuperscript{2}Meituan Longcat Team
 \textsuperscript{3}Peking University \\
 \textsuperscript{4}Xi'an Jiaotong University
 \textsuperscript{5}Nanjing University of Science and Technology
\\ \vspace{0.2cm}
 \texttt{pangxuan022@gmail.com},
 \texttt{zhangzijian14@meituan.com},
 \texttt{ninghliu@polyu.edu.hk}
}

\begin{document}
\maketitle
\begin{abstract}
Reinforcement Learning (RL) has enabled LLMs to excel in objective reasoning tasks such as mathematics and code generation.
However, applying RL to open-ended tasks, such as creative writing, remains challenging because LLM-as-a-judge reward models often exhibit stylistic biases and positional inconsistencies, leading to unstable supervision.
To address this, we propose OPERA (Objective Perplexity-based Reflective Alignment), which replaces unreliable external judges with intrinsic rewards derived from perplexity dynamics.
Specifically, we derive an intrinsic reward signal from perplexity dynamics, quantifying uncertainty reduction at critical reflective states.
During the cold-start phase, we introduce a data synthesis method that leverages carefully designed guiding words to generate diverse reasoning traces, along with perplexity-prioritized rollouts that utilize internal log-probabilities to identify logically consistent reasoning branches. This pipeline yields a large-scale dataset comprising 20,000 high-quality reasoning trajectories.
Empirical evaluations consistently demonstrate the scalability and efficacy of our approach in alignment for open-ended tasks.
Implementing OPERA on Qwen3-8B establishes a new state-of-the-art among open-source models, achieving parity with or surpassing proprietary models like Gemini2.5 and MiniMax-M2.5 in some open-ended tasks.
The code is available at \url{https://github.com/pangpang-xuan/OPERA}.
\end{abstract}

\section{Introduction}
Recent breakthroughs in LLMs have been driven by Reinforcement Learning with Verifiable Rewards (RLVR)~\cite{guo2025deepseek,team2025kimi,zhang2025survey}.
RLVR is particularly effective in logical with objective evaluation, such as mathematics and programming~\cite{zeng2025simplerl} where model outputs can be verified against binary feedback.
Reward-based training has been extended to creative writing, where evaluation is inherently open-ended.
Existing methods use pairwise writing supervision~\cite{Writing-Zero,li2026rewarding} and pairwise comparison rewards~\cite{lei2025writing,zhang2025extending,cao2026dpwriter,wu2025longwriter} to refine adaptive outputs. 
Other approaches adopt rubric-based rewards~\cite{gunjal2025rubrics} to decompose writing quality into interpretable dimensions, and provide structured, expert-aligned feedback. This design helps bridge the gap between binary correctness and coarse preference rankings.

Despite these improvements, the subjective reward mechanisms still faces major challenges.
A common issue is self-enhancement bias~\cite{ye2024justice}, where models favor responses that resemble their own stylistic preferences, such as specific structural patterns, rather than responses that are more creative or factually accurate.
Another limitation is positional bias~\cite{zheng2023judging}, where the judge systematically favors a response based on its position within the context window rather than its actual quality.
More fundamentally, LLM-based judges remain subjective and poorly calibrated, which limits their reliability as evaluation tools. 
These weaknesses make it difficult to build robust reward models for open-ended tasks. Establishing a more objective evaluation framework is therefore essential for advancing performance in open-ended domains.
\begin{figure*}[ht]
  \centering
  \includegraphics[width=0.88\textwidth]{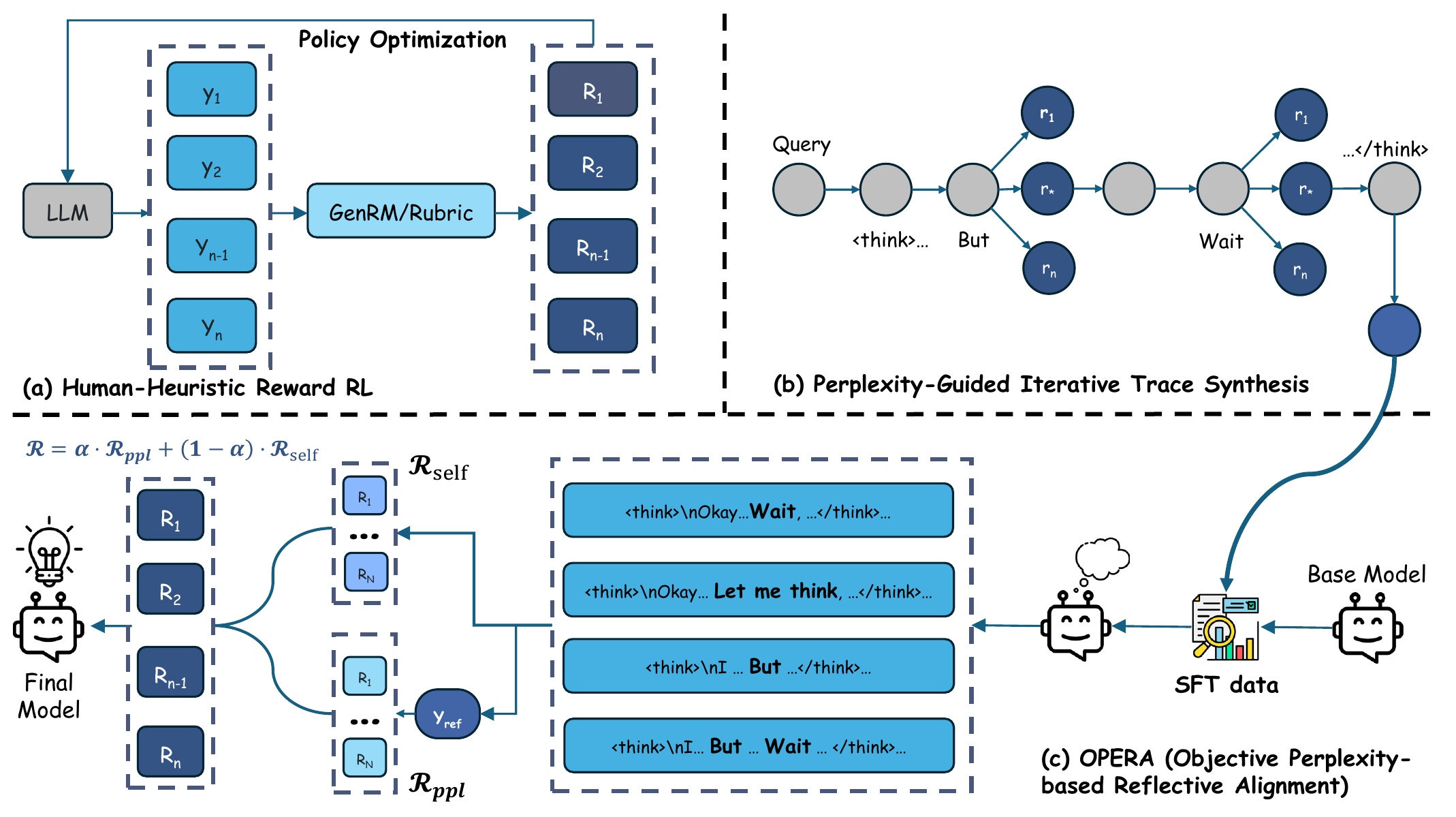}
  \caption{
  Overview of OPERA and cold-start reasoning trace synthesis. (a) Traditional reinforcement learning with LLM-based rewards. (b) Cold-start reasoning trace synthesis via perplexity-guided iterative generation. (c) OPERA (Objective Perplexity-based Reflective Alignment) for open-ended tasks.
  }
  \label{fig:abs}
\end{figure*}

To address these limitations, we introduce OPERA \textbf{(Objective Perplexity-based Reflective Alignment)}, a framework that bridges the gap between objective evaluation metrics and open-ended task performance, as illustrated in Figure~\ref{fig:abs}.
Unlike traditional approaches that rely on LLM-based judges, OPERA uses a more objective reward function grounded in perplexity dynamics.
More specifically, we propose a composite reward function that integrates local uncertainty reduction at reflective tokens with a global, group-relative reward, creating a robust reward signal especially suitable for domains where verifiable ground truth is inherently unavailable.
By measuring the differential change in PPL immediately before and after reflection tokens and optimizing for the intrinsic utility of its own deliberation, the model ensures that each reasoning step contributes to a more stable, high-likelihood response.
This approach effectively mitigates unstable reward by tokening the alignment process in the model’s internal log-probability, shifting the optimization target to the structural of the reasoning trajectory.
Consequently, OPERA fundamentally transitions reward functions from high-variance, LLM-based judgments to the internal logical consistency and predictive confidence of the model’s own reasoning.

During cold-start training, we propose a novel data synthesis pipeline called~\textbf{perplexity-guided iterative trace synthesis}, which shifts evaluation from external judgment to internal statistical consistency.
Our data synthesis is built upon two core components.
First, we introduce a cognitive braking interruption that leverages reflection tokens (e.g., “wait” or “but”) as heuristic indicators of cognitive conflict.
This mechanism prompts the model to pause its initial generation and engage in recursive System 2 deliberation~\cite{li2025system}, enabling a deeper and more careful reasoning process.
Second, we employ perplexity-prioritized rollouts, leveraging the model’s internal log-probabilities as an objective scoring metric to identify the most logically consistent reasoning branches.
This pipeline produces a large-scale dataset of 20,000 high-quality reasoning trajectories, which we use to perform SFT~\cite{ouyang2022training} as a cold start.

We evaluate the efficacy of OPERA across five benchmarks, demonstrating its scalability.
When applied to Llama-3.1-8B, OPERA boosts average benchmark performance by \textbf{125\%}, including a 22.54-point gain on WritingBench.
On Qwen3-8B, it sets a new open-source state-of-the-art and matches or even surpasses proprietary models like GPT-4o, Gemini2.5, and MiniMax-M2.5 on creative writing tasks.
These empirical results validate OPERA as a robust and scalable framework for open-ended reinforcement learning, providing a path toward high-fidelity reasoning without the need for external LLM-based judges.
Our contributions can be summarized as follows:
\begin{itemize}
    \item We present OPERA (Objective Perplexity-based Reflective Alignment), an objective reward function that leverages perplexity as a proxy for quality in open-ended tasks.
    \item In the cold-start training phase, we introduce Perplexity-Guided Iterative Trace Synthesis, which shifts evaluation from external validation to the model’s internal statistical consistency.
    \item We evaluate our method on two LLMs across five benchmarks and conduct detailed analyses to explain why perplexity serves as an effective proxy for human preferences.
\end{itemize}

\section{OPERA: Objective Perplexity-based Reflective Alignment}
\label{text:opera}
In contrast to objective tasks (e.g., maths or programming) where rewards are typically verifiable, open-end tasks lack an objective success criterion.
Existing approaches~\cite{Writing-Zero,li2026rewarding, zhang2026grad2reward} rely on LLM-based judges or heuristic rewards, which introduce bias and instability. 
We address this challenge by defining an intrinsic reward as a proxy, based on the model’s internal perplexity dynamics.
that quantifies the functional gain of internal reflection by utilizing perplexity (PPL)~\cite{han2025self} as a proxy for latent semantic quality.
This enables supervision of the latent strategy space, encouraging reasoning paths that proactively support error correction and stylistic alignment.

\subsection{Preliminary: Open-Ended Tasks Benefit from Reasoning Processes}
We conduct a preliminary experiment to demonstrate that the quality of reasoning influences the performance in open-ended tasks. 
Specifically, we replace the reasoning traces of a baseline model (Qwen3-8B) with those generated by stronger teacher models (DeepSeek, LongCat, and Qwen3-32B). 
Table~\ref{tab:performance-comparison} shows that this substitution consistently improves baseline model performance on creative writing benchmarks.
This result indicates that, \textbf{although reasoning traces are commonly used to improve performance in objective tasks such as maths and programming, they could also benefit open-ended tasks}.

Motivated by this observation, we propose improving reasoning as a proxy for enhancing open-ended generation. For LLMs operating in reasoning mode~\cite{guo2025deepseek,team2026longcat,qwen3} , the output typically contains two parts: a thought process (within <think>...</think> tags) and a final response.
In addition, we introduce a set of predefined \textbf{self-reflection tokens} $K$, such as “wait”, “but”, and “let me think”~\cite{wang2025reverse}, and encourage the model to generate them during reasoning.
These tokens provide an \textit{explicit signal} that the model is revising or reconsidering its current reasoning state. Without this constraint, improvements in log-probability may simply reflect normal token generation rather than genuine self-correction.

\begin{table}[t]
\centering
\caption{Performance on Arena Hard V2 Creative Writing. "Replace w/" denotes a stronger model's reasoning trace is injected into the prompt to guide the baseline model.}
\label{tab:performance-comparison}
\resizebox{0.45\textwidth}{!}{
    \begin{tabular}{lc}
    \toprule
    \textbf{Model} & \textbf{Arena Hard V2} \\
    \midrule
    Qwen3-8B & 25.7 \\
    \midrule
    Replace w/ DeepSeek~\cite{guo2025deepseek} & 26.3 \\
    Replace w/ LongCat~\cite{team2026longcat} & 31.8 \\
    Replace w/ Qwen3-32B~\cite{qwen3} & 27.6 \\
    \bottomrule
    \end{tabular}
}

\end{table}

\subsection{Self-Reflection Reward}
We introduce a self-reflection reward to encourage the model to reduce output uncertainty through the appropriate use of predefined self-reflection tokens.
Here, $x$ denotes the input prompt, $y_{ref}$ denotes the reference response, and $\langle \text{think} \rangle$ represents the opening token of the reasoning process. We define the baseline log-probability as $L_{base} = \log P(y_{ref} \mid x, \langle \text{think} \rangle)$
which measures the model’s initial expectation of the reference response before substantial reasoning occurs.
Then, to quantify internal reflection dynamics, we decompose the thinking process into sequential steps $S = \{s_1, s_2, \dots, s_n\}$.
At each step $s_j$, we compute the conditional log-probability:
$\log P_j = \log P(y_{ref} \mid x, s_{1:j})$, 
where $s_{1:j}$ denotes the reasoning trajectory up to step $j$.
A reflection step $s_j$ is considered productive only if it satisfies two conditions:
(1) it contains a predefined self-reflection token from the keyword set $K$;
(2) increases the log-probability of the reference response. 
The raw progress score $\mathcal{R}_{raw}$ is then defined as the cumulative count of productive reflections.
Formally:
\begin{equation}
    \mathcal{R}_{raw} = \sum_{j=1}^{n} \mathbb{I}\left((\log P_j - \log P_{j-1} > 0) \right),
\end{equation}
This formulation considers only whether a reflection step improves the target likelihood, rather than the magnitude of the improvement, which prevents reward hacking caused by local log-probability fluctuations.

To mitigate the risk of reward hacking by generating excessively long and redundant reasoning traces, we normalize the accumulated reward with a tangent function:
\begin{equation}
    \mathcal{R}_{self} = \tanh\left( \frac{\mathcal{R}_{raw}}{\tau} \right) ,
\end{equation}
where $\tau$ is a temperature hyperparameter that controls the saturation threshold.
This normalization provides strong incentives for the first few successful self-corrections while gradually diminishing rewards for additional reasoning steps. Consequently, the model learns to perform concise and meaningful self-correction instead of generating unnecessarily long reasoning traces.

\subsection{IGRP: In-Group Relative Perplexity Reward}
To evaluate the overall quality of both the reasoning trace and the final response, we further introduce the In-Group Relative Perplexity Reward (IGRP).
This metric follows the core principle of Group Relative Policy Optimization (GRPO)~\cite{guo2025deepseek}, which improves the policy through comparisons among multiple outputs generated from the same prompt $x$.
For a peer group of $N$ completions, we compute the joint log-probability of the reference response $y_{ref}$ conditioned on the full generated sequence: 
\begin{equation}
L_{hybrid}^i = \log P(y_{ref} \mid x, z^i, y^i),
\end{equation}
where $z^i$ represents the reasoning trace and $y^i$ the final response for the $i$-th completion.
We define the IGRP reward $\mathcal{R}_{ppl}$ as the normalized relative rank of a completion within its peer group. 
Specifically, for a given sample $x$ within a group of size $N$, the self-reflection reward $\mathcal{R}_{ppl}^i$ for the $i$-th completion is:
\begin{equation}
    \mathcal{R}_{ppl}^i = \frac{1}{N-1} \sum_{j=1, j \neq i}^{N} \mathbb{I}(L_{hybrid}^i > L_{hybrid}^j).
\end{equation}
where $\mathbb{I}(\cdot)$ denotes the indicator function.
This formulation quantifies the model's relative confidence by calculating the fraction of peer completions outperformed by the current sample, mapping certainty to a normalized reward score $\mathcal{R}_{ppl} \in [0, 1]$.
By utilizing a relative ranking rather than absolute log-probability values, the reward signal becomes less sensitive to variations in prompt difficulty.
The model is penalized only when its predictive likelihood is lower than the completions generated by its peers.
This formulation provides a stable optimization signal and encourages the model to produce efficient reasoning trajectories that lead to high-likelihood outputs.

\subsection{Hybrid Reward Function}
To improve open-ended reasoning while preserving performance on objective tasks, we introduce a hybrid reward function.
The overall reward function, $\mathcal{R}$, is formulated as a task-specific objective conditioned on the task domain $\mathcal{D} \in \{\mathcal{O}, \mathcal{E}\}$, where $\mathcal{O}$ and $\mathcal{E}$ respectively denote \textbf{objective} reasoning and \textbf{open-ended} reasoning.
\begin{equation}
\mathcal{R} = 
\begin{cases} 
\mathbb{I}(\text{parse}(y) = y_{gt}), & \text{if } \mathcal{D} = \mathcal{O}, \\
\alpha \cdot \mathcal{R}_{ppl} + (1 - \alpha) \cdot \mathcal{R}_{self}, & \text{if } \mathcal{D} = \mathcal{E} .
\end{cases}
\end{equation}

\subsubsection{Objective Reasoning}
For objective tasks such as math problems, we use binary rewards based on ground-truth accuracy.
Using a curriculum of verifiable math problems provides a stable guide during reinforcement learning.

\subsubsection{Open-ended Reasoning}
For open-ended domains such as creative writing, where a unique ground truth is absent, we utilize a weighted ensemble of $\mathcal{R}_{ppl}$ and $\mathcal{R}_{self}$.
It is designed to catalyze the emergence of autonomous self-correction behaviors by rewarding the model not only for the quality of the final output but for the utility of its internal reasoning trajectory.

\section{Perplexity-Guided Iterative Reasoning Trace Synthesis for Cold Start}
Directly applying RL to a base model often results in unstable optimization, reward hacking, and superficial alignment~\cite{wang2026reward, fu2025reward}.
To provide a stable initialization for RL training, we develop an \textbf{iterative synthesis method} for cold-start supervised fine-tuning.
The pipeline combines two components:
(1)~cognitive braking, which induces reflective reasoning; (2)~perplexity-prioritized rollouts, which select coherent continuations during search.

\subsection{Cognitive Braking}
To construct cold-start supervision traces with explicit reflective structure, we introduce a cognitive braking mechanism during reasoning generation.
Specifically, upon generating a predefined reflection token (e.g., “wait”, “let me think”), as introduced in Section~\ref{text:opera}, the model exits the current trajectory and revisits its intermediate reasoning before continuing.

Within our synthesis method, cognitive braking serves as the structural controller of the reasoning process.
It determines where reflective branching occurs and provides the starting states for subsequent candidate rollout generation.
As a result, the generated trajectories naturally contain explicit revision patterns and intermediate reconsideration behaviors, forming structured reasoning traces suitable for cold-start supervised fine-tuning.

\subsection{Perplexity-Prioritized Rollouts}
A key challenge in synthesizing cold-start rethinking data is maintaining the logical coherence during self-correction.
At each reflection token, the model is prompted to generate $k$ parallel candidate steps, $\mathcal{C} = \{c_1, c_2, \dots, c_k\}$, each extending the current reasoning trace. We must then select the continuation that preserves consistency.
For a candidate $c_i$, we construct the full sequence $X^{(i)} = \text{prompt} \oplus x_{<t} \oplus c_i$.
This candidate is then scored by perplexity (PPL), which averages log-likelihood of generating $X^{(i)}$, defined as:

\begin{equation}
\text{PPL}(c_i) = \exp (-\frac{1}{|X^{(i)}|} \sum_{t=1}^{|X^{(i)}|} \log P(X(i))),
\end{equation}

We then select the candidate with minimum perplexity: $c^* = \arg\min_{c_i \in \mathcal{C}} \text{PPL}(c_i)$.
By prioritizing paths with lower perplexity, we ensure that the synthetic trace represents the most statistically probable and thus logically consistent progression according to the model’s learned distribution.
This selection process effectively filters out noisy, low-confidence, or divergent rethinking steps, resulting in a high-quality trajectory for further fine-tuning.

The synthesis process proceeds recursively. Given a prompt $x$, the model alternates between generation and selection, extending the reasoning trace $y$ until it produces the terminal </think> token. The final response is then generated as $z=LLM(|x, y)$.
This procedure generates self-correcting reasoning trajectories that capture reflective revision processes and provide high-quality supervision signals for the SFT stage, enabling the model to learn robust reasoning and self-reflective behaviors.
However, since SFT alone is limited on open-ended tasks due to the gap between static supervision and dynamic exploration, we further apply RL to improve the model’s inference capabilities, reasoning adaptability, and generalization performance.

\section{Experiments}
\subsection{Experimental Setup}
\paragraph{Models.}
To evaluate the efficacy of our proposed method, we conduct extensive experiments using Llama3.1-8B~\cite{grattafiori2024llama} and Qwen3-8B~\cite{qwen3} as our base models.

\paragraph{Benchmarks.}
To ensure a comprehensive evaluation, we evaluate our method on five benchmarks: AlignBench~\cite{liu2024alignbench}, HelloBench~\cite{que2024hellobench}, EQ-Bench creative writing~\cite{paech2023eq}, WritingBench~\cite{wu2025writingbench}, and MATH500~\cite{hendrycks2021measuring}.
Our evaluation on AlignBench focuses on one primary domain: writing ability (\textbf{AB-W}).
These writing tasks are chosen to test not only language skills but also creativity. They require the model to produce more complex and expressive content, such as poems and fictional stories.
On HelloBench, we evaluate two primary domains: text completion (\textbf{HB-C}), which assesses the model's capacity for long-form generation, and heuristic text generation (\textbf{HB-G}), which focuses on content creation following specific stylistic or structural constraints.

\paragraph{Training Data.}
To construct the cold start SFT dataset, we used semantic clustering~\cite{kuhn2007semantic} to ensure diversity.
This process yielded a curated set of 20,000 raw entries filtered from LongWriter-6k~\cite{bai2024longwriter}, WildChat~\cite{zhao2024wildchat}, LitBench-Train~\cite{fein2026litbench} and OpenThought~\cite{guha2025openthoughts}.
We then utilized Qwen3-32B-Instruct~\cite{qwen3} as the generator model.
For the RL data, we subsampled GSM8K~\cite{cobbe2021training} and DeepWriting~\cite{wang2025reverse}. 
Detailed implementations are shown in Appendix~\ref{app:how_sft_data}.

\paragraph{Evaluation Protocols.}
Due to the inherent subjectivity of open-ended tasks, we follow established protocols by employing high-capacity LLMs as automated evaluators for our benchmarks.
While we recognize that this approach may introduce systemic biases, it currently provides the most scalable and consistent framework for assessing nuanced generative quality at scale.
On HelloBench, we apply the rescaling formula $S = (\text{score} - 0.75) \times 4$.

\paragraph{Baselines.}
Proprietary LLMs: GPT-4o-0513~\cite{hurst2024gpt}, Gemini 2.5-pro~\cite{comanici2025gemini}, and MiniMax-M2.5~\footnote{https://huggingface.co/MiniMaxAI/MiniMax-M2.5}.
Open-source LLMs:
LongWriter-8B~\cite{bai2024longwriter}, an open-source model optimized for ultra-long text generation;
DeepWriter-8B~\cite{wang2025reverse}, featuring iterative planning and self-reflection mechanisms; LongWriter-Zero-32B~\cite{wu2025longwriter}, a purely RL-based model capable of generating coherent passages.

\paragraph{Implementation Details.}
We employ ms-swift~\cite{zhao2025swift} as our SFT framework, training for 5 epochs with a learning rate of $1 \times 10^{-5}$.
For the RL phase, we utilize the verl~\cite{sheng2025hybridflow} framework to implement GRPO~\cite{shao2024deepseekmath}.
Training is conducted with a batch size of 64, with a response of 10,240 tokens.
We trained the actor model for one epoch using 32$\times$ NVIDIA H800 GPUs, using a learning rate of $1 \times 10^{-6}$ and 16 rollouts at a temperature of 1.0.
More details are detailed in Appendix~\ref{app:canshu}.

\subsection{Main Results}
The detailed experimental results in Table~\ref{tab:main_exp} reveal several key insights:
\paragraph{OPERA performs well on most tasks and models.}
We evaluated OPERA across two models, observing consistent and substantial improvements across four creative writing benchmarks and one mathematical task.
Qwen3-8B-OPERA significantly outperformed the strong open-source baseline in all categories.
This performance gap is most evident in the Creative Writing V3 benchmark, where Qwen3-8B-OPERA achieved an average improvement of over 10 points relative to DeepWriter-8B.
Furthermore, Llama3.1-8B-OPERA achieved an average score of 40.97, representing a \textbf{121.8}\% improvement over the base model's score of 18.47.
Notably, Llama3.1-8B-OPERA surpassed DeepWriter-8B and LongWriter-Zero-32B on HelloBench, suggesting that the OPERA objective effectively guides reinforcement learning toward superior reasoning and synthesis.
Furthermore, our results demonstrate that performance on math tasks remained uncompromised by the hybrid reward function.
This show that indicates that OPERA avoids the typical alignment tax, preserving core reasoning capabilities while simultaneously enhancing generalization across diverse domains.
We also extended our experiments to Qwen3-32B, as shown in Appendix~\ref{app:scaling_32b}.

\paragraph{OPERA performs comparable with state-of-the-art proprietary models.}
We evaluated our models against state-of-the-art proprietary models.
Across multiple creative writing benchmarks, Llama3.1-8B-OPERA demonstrates parity with or superiority over GPT-4o.
Notably, Llama3.1-8B-OPERA achieves a score of 30.08 on HB-C and 49.83 on HB-G, significantly outperforming GPT-4o-0513 (21.52 and 38.02, respectively).
On specialized benchmarks including Creative Writing V3 and WritingBench, Qwen3-8B-OPERA achieves performance competitive with Gemini-2.5-pro, effectively bridging the parameter gap between open-source models and proprietary frontier models.

\begin{table*}[t!]
  \centering
  \caption{Main performance comparison on creative writing and mathematical benchmarks. OPERA demonstrates competitive performance against leading proprietary models and significantly outperforms other open-source models.}
  \label{tab:main_exp}
  \resizebox{0.85\textwidth}{!}{
  \begin{tabular}{lcccccccc}
    \toprule
    \textbf{Model} & \textbf{AB-W} & \textbf{HB-C} & \textbf{HB-G} & \textbf{Creative Writing V3} & \textbf{WritingBench} & \textbf{MATH500} & \textbf{Avg}  \\
    \midrule
    \multicolumn{7}{l}{\textit{\textbf{Proprietary LLMs}}} \\ 
    \midrule
    GPT-4o-0513 &  6.37 & 21.52 & 38.02 & 61.76 & 66.19 & 78.00 & 45.31 \\
    Gemini-2.5-pro & 7.24 & 52.98 & 62.89 & 77.19 & 76.35 & 98.60 & 62.54 \\
    MiniMax-M2.5 & 6.84 & 26.80 & 59.33 & 80.80 & 80.39 & 98.40 & 58.76 \\
    \midrule
    \multicolumn{7}{l}{\textit{\textbf{Open-source LLMs}}} \\
    \midrule
    LongWriter-8B & 4.67 & -48.50 & -88.45 & 35.99 & 44.57 & 20.40 & -5.22 \\
    DeepWriter-8B & 6.63 & 26.33 & 0.25 & 62.22 & 73.70 & 80.40 & 41.59 \\
    LongWriter-Zero-32B & 6.95 & 21.26 & -55.48 & 51.11 & 77.44 & 73.40 & 29.11 \\
    \midrule
    \multicolumn{7}{l}{\textit{\textbf{Our Methods}}} \\
    \midrule
    Llama3.1-8B & 5.15 & 6.59 & -26.07 & 37.79 & 49.58 & 52.40 & 20.91 \\
    Llama3.1-8B-SFT & 5.89 & 29.68 & 37.53 & 63.21 & 71.30 & \textbf{58.20} & 44.30 \\
    Llama3.1-8B-OPERA & \textbf{6.15} & \textbf{30.08} & \textbf{49.83} & \textbf{67.09} & \textbf{72.12} & 56.80 & \textbf{47.01} \\
    \midrule
    Qwen3-8B & 6.64 & 35.77 & 37.12 & 65.25 & 75.08 & 95.60 & 52.58 \\
    Qwen3-8B-SFT & 7.00 & 38.70 & 41.96 & 71.92 & 76.63 & 94.80 & 55.17 \\
    Qwen3-8B-OPERA  & \textbf{7.13} & \textbf{38.91} & \textbf{49.79} & \textbf{72.89} & \textbf{78.19} & \textbf{96.20} & \textbf{57.19} \\
    \bottomrule
  \end{tabular}
}
\end{table*}

\subsection{Ablation Studies}
\paragraph{Perplexity-Guided Iterative Trace Synthesis.}
We set up four ablation experiments for data synthesis, and the specific settings can be shown in Appendix~\ref{app:ab} and the result as shown in Table~\ref{tab:ab_abs}.
Removing the synthesized data causes a huge drop in performance, most notably, a 43-point fall in Creative Writing V3 (71.92 $\rightarrow$ 28.64).
This shows that standard public datasets don’t provide enough structure for advanced reasoning.
Furthermore, we find that explicit reflection tokens and iterative search mechanisms are not merely structural;
they act as a workspace and a filter that helps the model think clearly, producing consistent and accurate results instead of mistakes.
We also reveal a key trade-off between search efficiency and peak performance.
Although limiting local search to the first five reflection tokens gives a fast and reliable baseline, the best performance is achieved only by expanding across all reflection tokens.
This shows that every step of deliberation adds unique value to the reasoning process.

\begin{table*}[t]
\centering
\caption{
Ablation study of OPERA. 
We compare the full model against variants with key components removed.
Numbers in parentheses denote the performance drop relative to the full OPERA model.
}
\label{tab:ab_abs}
\resizebox{0.92\textwidth}{!}{
\begin{tabular}{lcccc}
\toprule
\textbf{Model} 
& \textbf{HelloBench-C}
& \textbf{HelloBench-G}
& \textbf{Creative Writing V3}
& \textbf{WritingBench} \\
\midrule

Qwen3-8B-SFT
& 38.70
& 41.96
& 71.92
& 76.63 \\

\midrule

- w/o Synthesis Data
& 11.49 {\scriptsize($\downarrow$27.21)}
& -79.92 {\scriptsize($\downarrow$121.88)}
& 28.64 {\scriptsize($\downarrow$43.28)}
& 60.68 {\scriptsize($\downarrow$15.95)} \\

- w/o Iterative Search
& 34.20 {\scriptsize($\downarrow$4.50)}
& 42.69 {\scriptsize($\uparrow$0.73)}
& 67.77 {\scriptsize($\downarrow$4.15)}
& 70.55 {\scriptsize($\downarrow$6.08)} \\

- w/o Reflection Tokens
& 35.39 {\scriptsize($\downarrow$3.31)}
& 40.54 {\scriptsize($\downarrow$1.42)}
& 69.25 {\scriptsize($\downarrow$2.67)}
& 71.70 {\scriptsize($\downarrow$4.93)} \\

- Top-5 Token Rollout
& 36.22 {\scriptsize($\downarrow$2.48)}
& 45.93 {\scriptsize($\uparrow$3.97)}
& 69.84 {\scriptsize($\downarrow$2.08)}
& 70.78 {\scriptsize($\downarrow$5.58)} \\
\midrule
Qwen3-8B-OPERA
& 38.91
& 49.79
& 72.89
& 78.19 \\
\midrule
- w/o IGRP
& 33.69 {\scriptsize($\downarrow$5.22)}
& 46.60 {\scriptsize($\downarrow$3.19)}
& 68.56 {\scriptsize($\downarrow$4.33)}
& 77.01 {\scriptsize($\downarrow$1.18)} \\

- w/o Self-Reflection
& 38.92 {\scriptsize($\uparrow$0.01)}
& 41.63 {\scriptsize($\downarrow$8.16)}
& 71.29 {\scriptsize($\downarrow$1.60)}
& 77.15 {\scriptsize($\downarrow$1.04)} \\

\bottomrule
\end{tabular}
}
\end{table*}

\paragraph{Reward functions in OPERA.}
We also conducted an ablation study to test how sensitive OPERA is to the reward function, as summarized in Table~\ref{tab:ab_abs}, with detailed settings provided in Appendix~\ref{app:rl_ab}.
Relying solely on $\mathcal{R}_{self}$ leads to a performance regression across all metrics, notably a drop in Creative Writing V3 (72.89 $\rightarrow$ 68.56) suggesting that process-only rewards risk incentivizing superficial overthinking.
This confirms that while $\mathcal{R}_{self}$ successfully catalyzes self-correction, it requires an outcome-based global signal to ensure these cognitive efforts translate into tangible generative gains.
Conversely, isolating the IGRP reward triggers a sharp decline in HelloBench-G, indicating that outcome-based likelihoods alone are insufficient for tasks requiring complex creative synthesis.
This divergence highlights that while IGRP effectively optimizes the final response relative to the distribution, it lacks the fine-grained incentive required to navigate the nuanced, mid-course reasoning paths that $\mathcal{R}_{self}$ facilitates.

\section{Analysis and Discussions}
\subsection{Why OPERA can work?}
This section presents an analysis of the reward calculation in OPERA to achieve effective RL under objective perplexity, and more detail as shown in Appendix~\ref{app:how_align}.

In the self-reflection reward, the presence of a self-reflection token is essential because it provides an explicit and observable signal that the model has entered a reflective phase and attempted to revise or reconsider its reasoning process.
Without this constraint, improvements in log-probability could simply arise from ordinary token generation dynamics rather than genuine reflective behavior, making individual reflection steps difficult to identify or quantify.
By requiring self-reflection tokens, each step corresponds to a deliberate shift in reasoning strategy, enabling more consistent measurement and optimization of reflective capability.

A central challenge in open-ended reinforcement learning is designing an objective reward function that faithfully captures the preferences of a reward model.
To justify the use of IGRP as such an objective, we evaluate its alignment with high-capacity models, achieving a mean Kendall’s score of \textbf{54.01} and a mean Spearman’s score of \textbf{57.70}.
The Kendall~\cite{mcleod2005kendall} and Spearman~\cite{de2016comparing} correlations are particularly informative for open-ended generation tasks because they emphasize relative ranking quality rather than absolute likelihood.
Unlike standard perplexity, which is often biased by sequence length or surface-level fluency, IGRP evaluates responses relative to alternative generations under the same prompt, thereby isolating the logical contribution of the reasoning trace. The strong correlation with high-capacity evaluators suggests that minimizing conditional perplexity against expert references is well-aligned with semantic quality. These results support IGRP as a stable and objective reward proxy, effectively bridging raw likelihood optimization and human-aligned evaluation while avoiding the noise commonly associated with absolute perplexity metrics in non-deterministic generation tasks.

\subsection{Alignment: Perplexity vs. Judgment}
To carefully assess OPERA’s effectiveness compared to LLM-as-a-judge, we conduct a controlled study across two key stages of the alignment pipeline:
(i) cold-start SFT data curation, where an LLM-as-a-judge is used to select optimal reasoning steps;
(ii) reinforcement learning, where we use rubric-as-rewards as contrast our objective reward, the result as shown in Table~\ref{tab:llm_as_judge}.
\paragraph{Cold-start SFT data curation.}
Substituting PPL with an LLM-as-judge mechanism resulted in a performance regression across the evaluated benchmarks.
This consistent degradation suggests that while LLM judges are capable of assessing high-level semantic correctness, they are often blind to the distributional consistency required for a model to internalize a stable reasoning policy.
In contrast, our PPL-guided approach finds trajectories that naturally align with the model’s internal probability structure.
This alignment speeds up training and improves generalization, especially for long-generation tasks typical in open-ended writing.

\begin{table}
\centering
\caption{Comparative Analysis of Perplexity Guidance vs. LLM-as-a-Judge across SFT and RL Pipelines.}
\label{tab:llm_as_judge}
\resizebox{\linewidth}{!}{
\begin{tabular}{lcccc}
\toprule
\textbf{Model} & \textbf{HB-C} & \textbf{HB-G} & \textbf{Creative Writing V3} & \textbf{WritingBench} \\
\midrule
Qwen3-8B-SFT & \textbf{38.70} & 41.96 & \textbf{71.92} & \textbf{76.63} \\
- LLM-as-judge & 37.34 & \textbf{43.76} & 65.79 & 72.12 \\
\midrule
Qwen3-8B-OPERA & \textbf{38.91} & \textbf{49.79} & \textbf{72.89} & \textbf{78.19} \\
- Rubric as Reward & 32.52 & 45.06 & 71.47 & 76.40 \\
\bottomrule
\end{tabular}
}
\end{table}

\paragraph{Rubric as Rewards.}
Within the RL, the rubric-as-reward baseline underperforms the OPERA framework by a significant margin.
This disparity underscores a fundamental limitation of discrete, LLM-based evaluations:
they often provide a coarse-grained signal that lacks the resolution required to guide a model through complex, long-horizon reasoning trajectories.
In contrast, our proposed reward mechanism offers a more continuous and nuanced signal, effectively preserving reasoning depth by rewarding the structural and statistical integrity of each cognitive step.

\section{Related Works}
\subsection{Test-Time Computation}
Recent work has shown that increasing test-time computation can substantially improve language model performance.
Early progress was driven by Chain-of-Thought (CoT) prompting~\cite{wei2022chain}, which encourages models to generate intermediate reasoning steps for more accurate problem solving.
Building on CoT, Tree-of-Thoughts~\cite{yao2023tree} extends linear reasoning into a multi-branch search process, enabling models to explore, evaluate, and revise multiple reasoning paths.
More recently, organizations such as DeepSeek AI~\cite{guo2025deepseek} and OpenAI~\cite{jaech2024openai} have further demonstrated the effectiveness of scaling test-time reasoning for improving model capability.

\subsection{Reinforcement Learning}
Reinforcement Learning~\cite{wen2025reinforcement} has become a key paradigm for aligning LLMs with objective correctness.
Unlike RLHF~\cite{ouyang2022training}, RLVR relies on deterministic verifiers, such as answer matching or unit tests, to provide sparse but reliable feedback signals.
Recent advances, including DeepSeek-R1~\cite{guo2025deepseek}, demonstrate that RLVR can effectively induce long chain-of-thought reasoning~\cite{wei2022chain}. More recent studies have further extended RLVR beyond deterministic domains, with methods such as Writing-Zero~\cite{Writing-Zero} exploring self-principled rewards and reference-based matching for creative and open-ended generation tasks.

\section{Conclusion}
We introduce OPERA \textbf{(Objective Perplexity-based Reflective Alignment)}, a framework that shifts the alignment of open-ended reasoning from fallible external LLM-based judge toward intrinsic perplexity dynamics.
We derive a composite reward that combines an intrinsic self-reflection signal quantifying uncertainty reduction at reflective tokens with a reward based on relative predictive confidence.
During cold-start training, we introduce Perplexity-Guided Iterative Trace Synthesis, which leverages cognitive braking to trigger System 2 deliberation, uses perplexity-prioritized rollouts to ensure structural consistency, and generates 20,000 high-quality reasoning trajectories.
Overall, our method provides a scalable reinforcement learning objective for open-ended tasks where verifiable ground-truth feedback is not available.

\section*{Limitations}
While we acknowledge that LLM-based evaluation may introduce inherent model-specific biases, it remains a robust and widely adopted framework for scalable benchmarking.
Furthermore, although this current study primarily focuses on writing tasks, future iterations will extend to open-ended QA and multi-turn dialogue
These expansions will provide a more comprehensive assessment of the framework’s generative versatility and its performance in dynamic conversational contexts.



\bibliography{custom}

\appendix

\section{Influence of Reasoning Traces on Generative Performance}
\label{app:reasoning_trace}
To isolate the impact of thought process on task performance, we conducted a preliminary ablation study where the thought process of baseline LRMs were replaced with those generated by more capable teacher models.
The result as shown in Table~\ref{tab:performance-comparison_app}.
This demonstrates that, the thought process of an LRM is crucial to the quality of the final output; a higher quality thought process corresponds to a higher quality final output.
Therefore, for creative tasks, our goal is to use SFT or RL to improve the quality of the model's reasoning process, shifting the focus from enhancing the quality of the final output to enhancing the thought process itself.
\begin{table*}[htbp]
  \centering
  \small
  \caption{Performance comparison on AIME 2025 and Arena Hard V2. \textit{Replace} denotes a configuration where a high-capacity LRM's reasoning trace is injected into the prompt to guide the baseline model's response.}
  \label{tab:performance-comparison_app}
  \begin{tabular}{lccc}
    \toprule
    \textbf{Model} & \textbf{AIME 2025} & \multicolumn{2}{c}{\textbf{Arena Hard V2 (Score)}} \\
    \cmidrule(lr){3-4}
    & \textbf{(Pass@1)} & \textbf{Hard Prompt (G/P)\textsuperscript{$\dagger$}} & \textbf{Creative Writing} \\
    \midrule
    Qwen3-8B & 0.67 & 23.3 / 18.7 & 25.7 \\
    \midrule
    Replace DeepSeek~\cite{guo2025deepseek} & 0.80 & 51.7 / 46.9 & 26.3 \\
    Replace LongCat~\cite{team2026longcat} & \textbf{0.83} & \textbf{51.3 / 50.6} & \textbf{31.8} \\
    Replace Qwen3-32B~\cite{qwen3} & 0.76 & 46.4 / 44.0 & 27.6 \\
    \bottomrule
    \multicolumn{4}{l}{\textsuperscript{$\dagger$}\footnotesize G/P refers to Gemini and GPT as judges, respectively.}
  \end{tabular}
\end{table*}

\section{Training Data}
\label{app:data}

\subsection{Data Curation}
\label{app:how_sft_data}
To construct the cold-start SFT dataset, we employed semantic clustering~\cite{kuhn2007semantic} to ensure broad task diversity, followed by stratified random sampling across the identified categories.
Specifically, we utilized the BGE-M3~\cite{chen2024bge} to generate high-dimensional embeddings for the initial prompt pool. 
After clustering these embeddings, we performed proportional sampling from each cluster to maintain the original distribution while curating a representative subset.
The dataset was synthesized by filtering and aggregating high-quality examples from a diverse array of established sources, specifically LongWriter-6k~\cite{bai2024longwriter}, WildChat~\cite{zhao2024wildchat}, LitBench-Train~\cite{fein2026litbench} and OpenThought~\cite{guha2025openthoughts}.
This balanced distribution ensures the model develops both the creative linguistic fluidity required for long-form generation and the rigorous logical precision necessary for multi-step mathematical problem-solving.

To curate high-quality training samples for both mathematical reasoning and open-ended generation in RL phase, we performed a filtering process.
For our mathematical training set, we selectively retained only those prompts where the model achieved a correct solution in exactly \textbf{one} out of \textbf{eight} independent rollouts.
For creative writing tasks, we utilized an LLM-as-a-judge framework to evaluate the model's output against a gold-standard reference, employing a scoring scale of 0 to 5.
We specifically targeted samples with scores in the \textbf{2–3} range to facilitate effective error-correction training.
This rigorous selection process ultimately yielded a high-quality training corpus consisting of 1,165 writing entries and 509 mathematical reasoning samples.

\subsection{Differences with other methods}
\begin{figure*}[t!]
  \centering
  \includegraphics[width=0.85\linewidth]{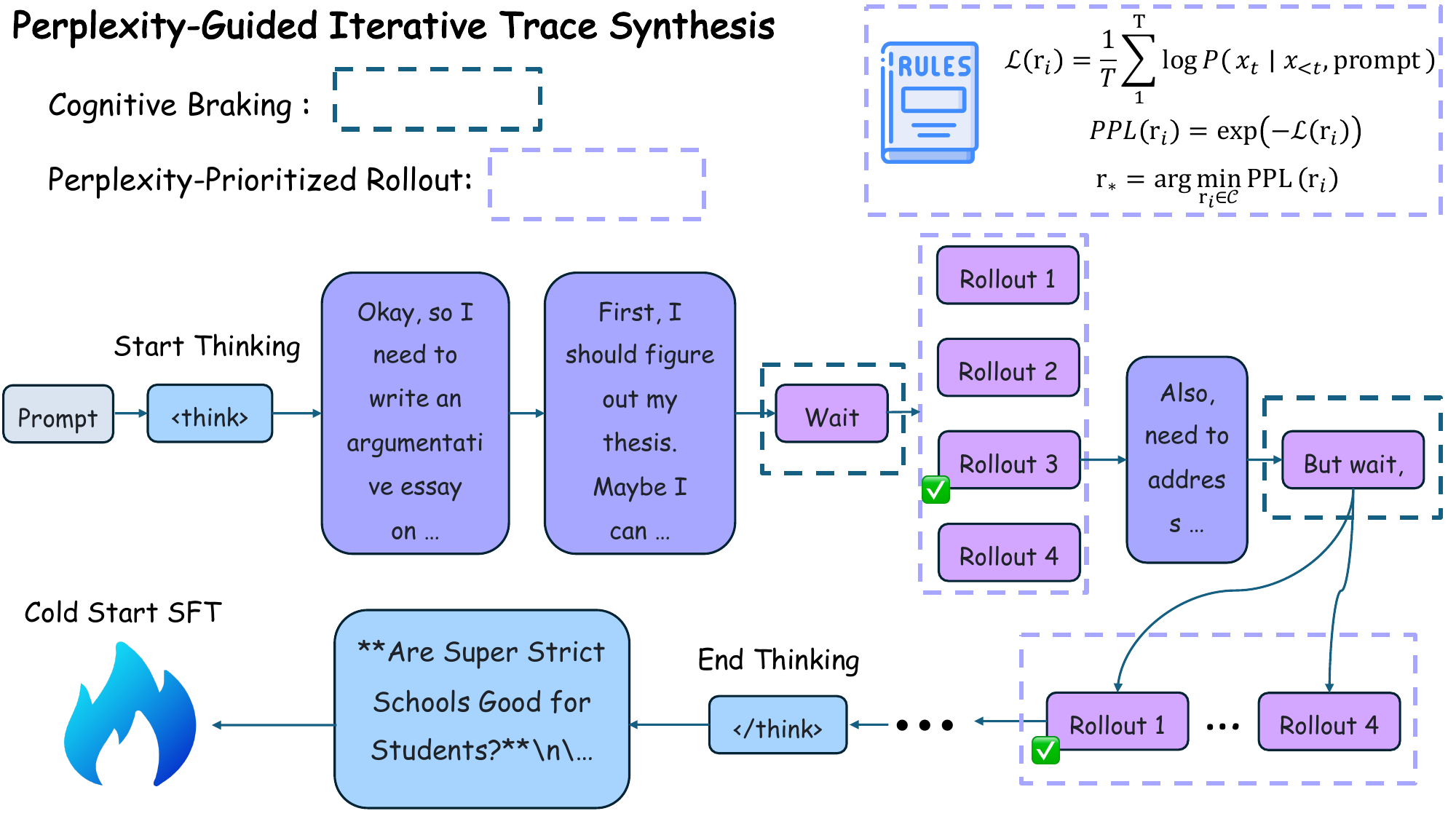}
  \caption{The overview of Perplexity-Guided Iterative Trace Synthesis in Cold Start SFT.}
  \label{fig:framework}
\end{figure*}
It is important to distinguish our iterative trace synthesis from methods like Monte Carlo Tree Search (MCTS)~\cite{zhang2024rest} and REER~\cite{wang2025reverse}.
Unlike REER, which performs local surgery on an existing, fixed trajectory to minimize the perplexity of a ground-truth answer, our method grows a trajectory autoregressively from left to right.
As illustrated in Figure~\ref{fig:framework}, the approach employs an asynchronous interruption mechanism in which designated reflection tokens serve as triggers for System 2 deliberation, enabling the model to adaptively allocate additional computational resources only when cognitive conflict is detected.
While MCTS relies on complex process reward models to navigate a global search tree, our method prioritizes internal logical consistency by selecting parallel candidate rollouts that minimize the model's own log-probabilities.
This recursive process results in a high-density, self-correcting gold standard trajectory that captures the meta-cognitive process of error detection and rectification, providing a more efficient and inspired foundation for subsequent fine-tuning than either static refinement or exhaustive tree search.

\subsection{Global PPL Landscape}
\begin{figure*}[ht]
  \centering
  \includegraphics[width=\textwidth]{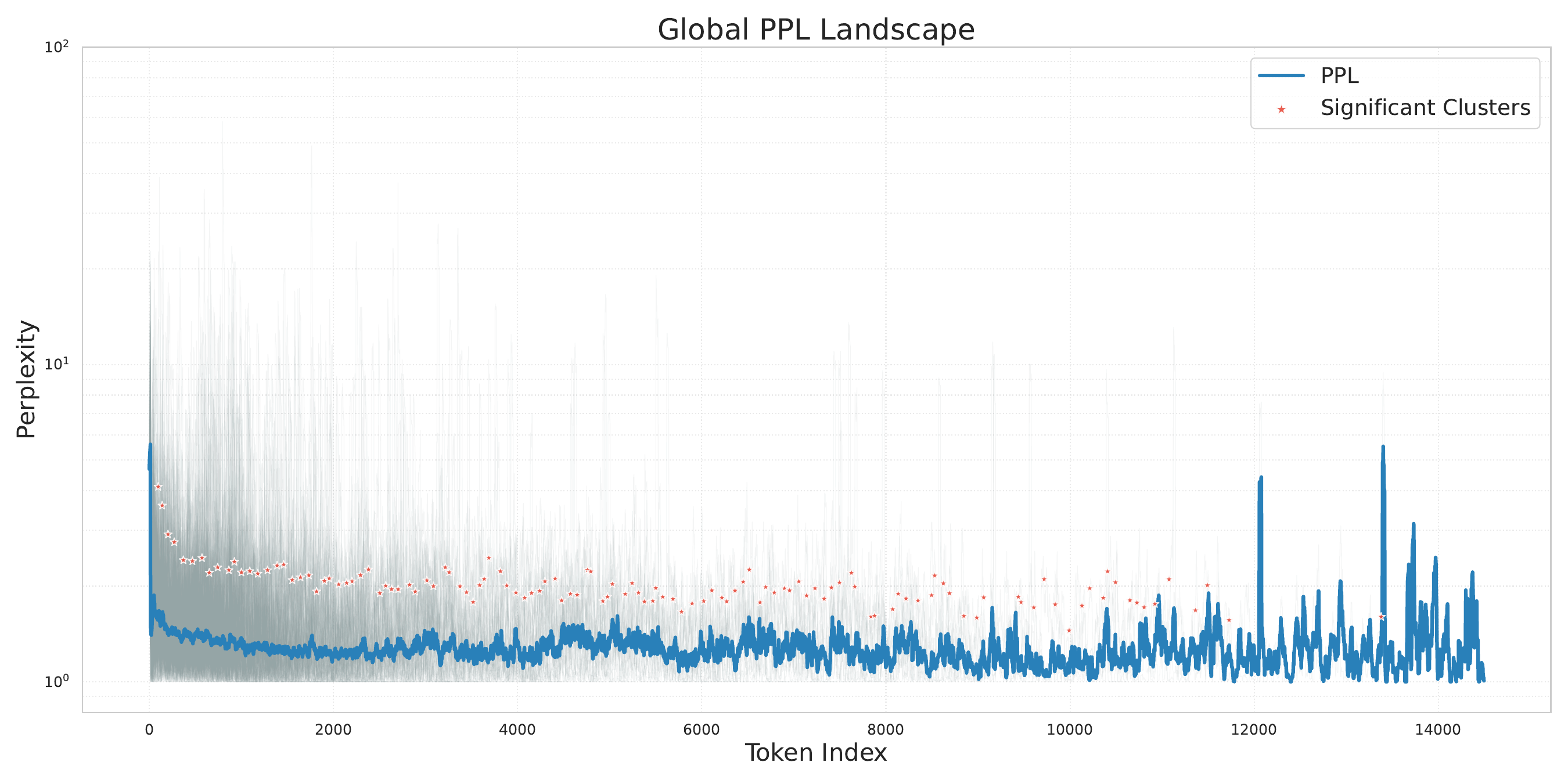}
  \caption{Global PPL Landscape of Reasoning Traces.}
  \label{fig:app_ppl}
\end{figure*}

We employ a sliding-window perplexity analysis to monitor the model's internal confidence during the chain-of-thought generation.
Using a window size of $W=25$, we calculate the local PPL to capture the dynamic shifts in the model's transition probabilities.
To identify critical reasoning junctions, we define significant clusters as tokens that:
(1) belong to a predefined set of reflexive keywords (\textit{e.g.}, 'but', 'wait'), and (2) exhibit a local PPL exceeding the global median by a threshold of $\tau = 1.15$, the result as shown in Figure~\ref{fig:app_ppl}.
The blue curve representing the median PPL remains remarkably stable and low throughout the majority of the reasoning process.

This indicates that the model maintains high structural confidence across most tokens, a hallmark of successful SFT on the dataset.
Moreover, we observe that as the token index increases (particularly beyond 12,500 tokens), the variance in local PPL begins to expand.
This visualizes the accumulation of uncertainty in ultra-long reasoning chains, where minor logical drifts can lead to significant predictive entropy.
A high density of clusters is observed in the initial reasoning phase (tokens 0–2,500). 
This suggests the model frequently engages in cognitive braking to calibrate its initial logic path before proceeding to stable derivation.
Isolated clusters appearing in the middle of a stable trajectory (\textit{e.g.}, around index 10,000 and 13,000) serve as objective signals for over-trust Penalty.
These points indicate where the model successfully identified a potential reasoning error and attempted a linguistic re-alignment.

\subsection{Analysis of Perplexity-Guided Iterative Trace Synthesis}
\begin{figure*}[ht]
  \centering
  \includegraphics[width=\textwidth]{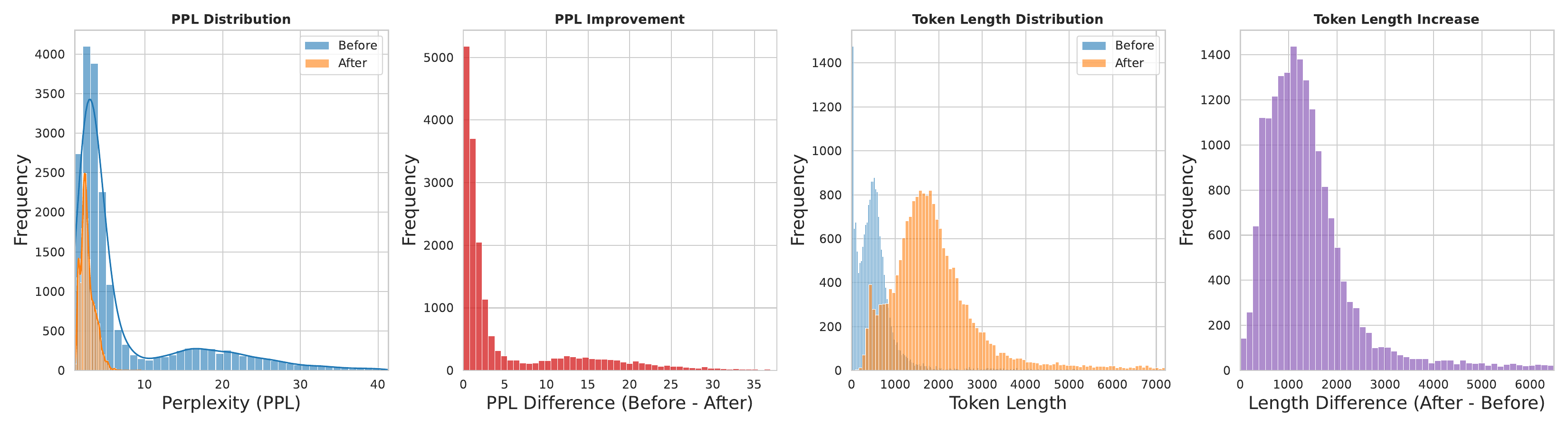}
  \caption{Analysis of Token Length \& Perplexity Before and After the Synthesis.}
  \label{fig:app_sft_analysis}
\end{figure*}

Analysis of the synthesis process, as illustrated in Figure~\ref{fig:app_sft_analysis}, confirms its empirical effectiveness.
Following the synthesis stage, the perplexity distribution exhibits a significant downward shift, with a vast majority of samples demonstrating marked improvement in PPL.
Concurrently, we observe a systematic increase in the token length of the reasoning trajectories.
This trend indicates that the synthesis process successfully expands initial, skeletal plans into more detailed and elaborate reasoning chains, effectively bridging the gap between high-level intent and granular logical execution.

\section{More detail in experiment}
\subsection{Implementation Details}
\label{app:canshu}
We implement the SFT phase using the ms-swift framework~\cite{zhao2025swift}, training the model for 5 epochs with a learning rate of $1 \times 10^{-5}$ and a warmup ratio of 0.05.
The training was conducted on a cluster of 8$\times$ NVIDIA H800 GPUs.
We utilized a per-device batch size of 1 with gradient accumulation set to 2, resulting in a total training duration of approximately \textbf{8 hours and 45 minutes}.
For the reinforcement learning phase, we utilize the Verl~\cite{sheng2025hybridflow} to implement GRPO~\cite{shao2024deepseekmath}.
The configuration utilized a factor of $\alpha = 0.3$ in reward function and a setting of $\tau = 2$. 
Training was executed on a high-performance cluster comprising 32$\times$ NVIDIA H800 GPUs. To ensure stability during the initial phases of optimization, the actor-model utilized a learning rate of $1 \times 10^{-6}$ paired with a warmup ratio of 0.4.
To manage the high dimensionality of long-form reasoning, we configured the system to support a maximum prompt length of 6,144 tokens and a generated response length of 10,240 tokens.
We utilized 16 rollouts per prompt at a temperature of 1.0, with a KL-divergence coefficient of 0.001 to ensure policy stability.
To optimize memory throughput, we applied a tensor model parallelism size of 2 for the rollouts.
The final model was trained for one epoch over \textbf{26} steps, with a total wall-clock time of approximately \textbf{6 hours and 10 minutes}.

\subsection{Evaluation Protocols}
To address the inherent subjectivity of open-ended generation, we adopted the established protocol of utilizing frontier Large Language Models as automated judges across our evaluation benchmarks.
\textbf{While we acknowledge the potential for model-specific biases in this paradigm, it currently represents the most scalable and consistent methodology for quantifying nuanced generative quality.}

For AlignBench and HelloBench, we employed GPT-4o (2024-08-06) as the primary evaluator.
For WritingBench, we utilized Claude-3.7 to capture the stylistic intricacies of the outputs. In Creative Writing V3, we implemented a dual-judge ensemble consisting of Gemini-2.5-Pro and GPT-4.1; scores were derived via uniform sampling and weighted averaging to mitigate individual model variance.
Finally, to enhance the resolution of comparative performance on HelloBench, we applied a linear rescaling formula: $S = (\text{score} - 0.75) \times 4$. This transformation maps the original outputs to a standardized range of $[-300, 100]$, effectively amplifying the delta between high-performing models.

During the inference phase, for the open-ended and creative benchmarks including AlignBench, Creative Writing V3, and WritingBench we utilized a temperature of 0.7.
For HelloBench, we use a temperature of 0.6.
Conversely, for the MATH500 benchmark, we employed greedy decoding (temperature = 0) to ensure deterministic reasoning paths for verifiable mathematical problems.

For our comparative evaluation, we selected three prominent open-source baselines: LongWriter, DeepWriter, and LongWriter-Zero-32b.
We utilized the official LongWriter-Llama-3.1-8B~\footnote{https://huggingface.co/zai-org/LongWriter-llama3.1-8b} and LongWriter-Zero-32B~\footnote{https://huggingface.co/THU-KEG/LongWriter-Zero-32B}.
In contrast, DeepWriter only released the training corpus rather than a pre-trained model, we independently trained a variant using their provided dataset and their implementation parameter to ensure a consistent experimental environment.
All models were then subjected to the same evaluation protocol as OPERA to facilitate a rigorous and direct performance comparison.

\subsection{Experiment Results}
To establish the statistical significance of our findings, we report the variance across our experimentals.
These metrics, provided in Table~\ref{tab:full_main_exp}, demonstrate the stability of the OPERA framework and ensure that the observed performance gains are consistent and reproducible.

\begin{table*}[htbp]
  \centering
  \large
  \caption{Main Performance Comparison (Mean $\pm$ Standard Deviation) on Creative Writing and Mathematical Benchmarks}
  \label{tab:full_main_exp}

  \resizebox{\textwidth}{!}{
  \begin{tabular}{lcccccc}
    \toprule
    \textbf{Model} & \textbf{AB-W} & \textbf{HB-C} & \textbf{HB-G} & \textbf{Creative Writing V3} & \textbf{WritingBench} & \textbf{MATH500} \\
    \midrule

    \multicolumn{7}{l}{\textit{\textbf{Proprietary LLMs}}} \\
    \midrule

    GPT-4o-0513 & $6.37 \pm 0.00$ & $21.52 \pm 6.62$ & $38.02 \pm 10.10$ & $61.76 \pm 3.44$ & $66.19 \pm 0.51$ & $78.00 \pm 0.00$ \\

    Gemini-2.5-pro & $7.24 \pm 0.00$ & $52.98 \pm 6.73$ & $62.89 \pm 8.56$ & $77.19 \pm 1.62$ & $76.35 \pm 0.71$ & $98.60 \pm 0.00$ \\

    MiniMax-M2.5 & $6.84 \pm 0.00$ & $26.80 \pm 9.83$ & $59.33 \pm 9.01$ & $80.80 \pm 1.75$ & $80.39 \pm 0.68$ & $98.40 \pm 0.00$ \\

    \midrule

    \multicolumn{7}{l}{\textit{\textbf{Open-source LLMs}}} \\
    \midrule

    LongWriter-8B & $4.67 \pm 0.00$ & $-48.50 \pm 20.43$ & $-88.45 \pm 27.94$ & $35.99 \pm 4.64$ & $44.57 \pm 1.16$ & $20.40 \pm 0.00$ \\

    DeepWriter-8B & $6.63 \pm 0.00$ & $26.33 \pm 8.84$ & $0.25 \pm 21.29$ & $62.22 \pm 3.20$ & $73.70 \pm 0.67$ & $80.40 \pm 0.00$ \\

    LongWriter-Zero-32B & $6.95 \pm 0.00$ & $21.26 \pm 10.52$ & $-55.48 \pm 22.99$ & $51.11 \pm 4.75$ & $77.44 \pm 0.73$ & $73.40 \pm 0.00$ \\

    \midrule

    \multicolumn{7}{l}{\textit{\textbf{Our Methods}}} \\
    \midrule

    Llama3.1-8B & $5.15 \pm 0.00$ & $6.59 \pm 8.10$ & $-26.07 \pm 16.48$ & $37.79 \pm 4.15$ & $49.58 \pm 0.80$ & $52.40 \pm 0.00$ \\

    Llama3.1-8B-SFT & $5.89 \pm 0.00$ & $29.68 \pm 8.37$ & $37.53 \pm 9.58$ & $63.21 \pm 3.75$ & $71.30 \pm 0.72$ & $\textbf{58.20} \pm 0.00$ \\

    Llama3.1-8B-OPERA & $\textbf{6.15} \pm 0.00$ & $\textbf{30.08} \pm 7.67$ & $\textbf{49.83} \pm 10.91$ & $\textbf{67.09} \pm 3.32$ & $\textbf{72.12} \pm 0.74$ & $56.80 \pm 0.00$ \\

    \midrule

    Qwen3-8B & $6.64 \pm 0.00$ & $35.77 \pm 6.62$ & $37.12 \pm 9.73$ & $65.25 \pm 3.11$ & $75.08 \pm 0.60$ & $95.60 \pm 0.00$ \\

    Qwen3-8B-SFT & $7.00 \pm 0.00$ & $38.70 \pm 7.24$ & $41.96 \pm 11.39$ & $71.92 \pm 2.25$ & $76.63 \pm 0.57$ & $94.80 \pm 0.00$ \\

    Qwen3-8B-OPERA & $\textbf{7.13} \pm 0.00$ & $\textbf{38.91} \pm 6.58$ & $\textbf{49.79} \pm 9.67$ & $\textbf{72.89} \pm 1.75$ & $\textbf{78.19} \pm 0.58$ & $\textbf{96.20} \pm 0.00$ \\

    \bottomrule
  \end{tabular}
  }

\end{table*}

\subsection{Ablation Studies}
We provides a detailed introduction and analysis under different ablation settings.

\subsubsection{Perplexity-Guided Iterative Trace Synthesis}
\label{app:ab}
\begin{itemize}
\item \textbf{Remove Synthesis Data}:
Removing our synthesized trajectories and training only on public datasets.
We observed that applying public datasets directly to SFT resulted in a precipitous performance degradation.
Specifically, in the Creative Writing V3 benchmark, performance plummeted by nearly \textbf{43} points (71.92 $\rightarrow$ 28.64), with similarly drastic declines observed in Hellobench.
These results validate a central hypothesis of this work: explicit reflection tokens and structured reasoning trajectories are not merely decorative, but essential catalysts for enhancing the model's downstream capabilities.

\item \textbf{Remove Iterative Search}:
During the generation of sequences containing reflective tokens, we bypass the rollout-based selection mechanism.
Specifically, we observe that reflection alone is insufficient;
if the specific step at which reflection occurs is sub-optimal, model capability diminishes significantly.
This finding validates the efficacy of perplexity-prioritized rollouts, demonstrating that the discovery of superior reasoning paths is essential for translating latent thinking into concrete generative gains.

\item \textbf{Remove Reflection Tokens}:
When these linguistic markers were removed maintaining only the local search, model performance consistently degraded across open-ended benchmarks (\textit{e.g.}, WritingBench: 76.63 $\rightarrow$ 71.70).
This suggests that explicit tokens for cognitive exploration and self-correction are not merely structural artifacts; rather, they provide a critical representational workspace that facilitates the creative divergence necessary for complex, artistic writing tasks.

\item \textbf{Top-5 Token Rollouts}:
While restricting expansion to the first five reflection tokens yields significant efficiency gains, our experiments demonstrate that comprehensive expansion across all reflection tokens provides the superior performance ceiling.
This indicates that while the perplexity metric is highly precise in the early stages of reasoning, the cumulative effect of local search across every cognitive marker is essential for reaching peak model capability.
This finding suggests that each stage of deliberation contributes unique structural value to the trajectory.

\end{itemize}

\begin{table*}[htbp]
  \centering
  \caption{Main Performance Comparison (Mean ± Variance) of ablation studies in Perplexity-Guided Iterative Trace Synthesis.}
  \label{tab:ab_sft_full}
  \resizebox{\textwidth}{!}{
  \begin{tabular}{lcccc}
    \toprule
    \textbf{Model} & \textbf{HelloBench-C} & \textbf{HelloBench-G} & \textbf{Creative Writing V3} & \textbf{WritingBench}\\
    \midrule
    Qwen3-8B-SFT & $\textbf{38.70} \pm 7.24$ & $41.96 \pm 11.39$ & $\textbf{71.92} \pm 2.25$ & $\textbf{76.63} \pm 0.57$ \\
    \midrule
    - Remove Synthesis Data & $11.49 \pm 7.80$ & $-79.92 \pm 27.06$ & $28.64 \pm 4.42$ & $60.68 \pm 0.82$ \\
    - Remove Iterative Search & $34.20 \pm 6.89$ & $42.69 \pm 12.61$ & $67.77 \pm 3.62$ & $70.55 \pm 0.63$ \\
    - Remove Reflection Tokens & $35.39 \pm 8.78$ & $40.54 \pm 9.42$ & $69.25 \pm 3.56$ & $71.70 \pm 0.67$ \\
    - Top5-token Rollout & $36.22 \pm 7.19$ & $\textbf{45.93} \pm 12.18$ & $69.84 \pm 2.90$ & $70.78 \pm 0.61$ \\
    \bottomrule
  \end{tabular}
}
\end{table*}

\subsubsection{Reward functions in OPERA}
\label{app:rl_ab}
\begin{itemize}
\item \textbf{Only Self-Reflection Reward}:
The model experiences a significant performance decline across all metrics when relying solely on the self-reflection reward, with particularly sharp drops in Creative Writing V3 ($72.89 \rightarrow 68.56$).
Because the self-reflection reward focuses primarily on the reasoning process specifically internal pivots and keyword triggers, it risks incentivizing the model to overthink without ensuring that this latent effort translates into a superior final response.
This suggests that while $\mathcal{R}_{ref}$ successfully catalyzes self-correction behaviors, it requires a grounding mechanism like IGRP to provide a global quality signal. Without IGRP, the model may optimize for the appearance of reasoning while failing to achieve the semantic alignment necessary for high-quality creative and constrained generation.

\item \textbf{Only IGRP}:
The removal of the self-reflection reward leads to a sharp decline in HelloBench-G performance ($49.79 \rightarrow 41.63$), while other benchmarks remain stable or show marginal improvements.
This divergence suggests that while the IGRP reward effectively optimizes the final outcome's likelihood relative to peers, it is insufficient for tasks requiring complex creative synthesis.
The drop in performance indicates that without the explicit incentive provided by $\mathcal{R}_{ref}$ to catalyze mid-course corrections, the model struggles to navigate the nuanced reasoning paths necessary for high-level creative generation.
\end{itemize}

\begin{table*}[htbp] 
\centering
\caption{Main Performance Comparison (Mean ± Variance) of ablation studies in OPERA reward functions.}
\label{tab:ab_rl_full}
\resizebox{\textwidth}{!}{
\begin{tabular}{lcccc}
\toprule \textbf{Model} & \textbf{HelloBench-C} & \textbf{HelloBench-G} & \textbf{Creative Writing V3} & \textbf{WritingBench}\\
\midrule
Qwen3-8B-OPERA & $38.91 \pm 6.58$ & $\textbf{49.79} \pm 9.67$ & $\textbf{72.89} \pm 1.75$ & $\textbf{78.19} \pm 0.58$ \\
\midrule
- w/o IGRP & $33.69 \pm 7.17$ & $46.60 \pm 12.80$ & $68.56 \pm 3.53$ & $77.01 \pm 0.52$ \\
- w/o self-reflection & $\textbf{38.92} \pm 7.18$ & $41.63 \pm 10.68$ & $71.29 \pm 2.15$ & $77.15 \pm 0.57$ \\
\bottomrule
\end{tabular}
}
\end{table*}

\subsubsection{Without SFT phase}
We observed that omitting the SFT phase and performing OPERA directly on the base model led to a significant degradation in performance.
This is likely attributable to the cold-start problem in latent strategy exploration. Without a supervised initialization to align the model with the reasoning syntax, the RL algorithm faces a sparse reward landscape.
The resulting gradients, derived largely from low-quality rollouts, induce distributional instability and catastrophic forgetting of pre-trained capabilities.
This underscores the necessity of SFT as a structural prior that constrains the search space to a regime of high-fidelity reasoning.

\begin{table*}[htbp] 
\centering
\caption{Ablation Analysis of the SFT-RL Pipeline. Performance benchmarks for models trained via direct Reinforcement Learning on the base model versus the standard SFT+RL pipeline. The "Base + RL" configuration demonstrates a performance collapse, highlighting the "cold start" challenge in autonomous reasoning discovery.}
\label{tab:ab_rl}
\resizebox{0.95\textwidth}{!}{
\begin{tabular}{lcccc}
\toprule \textbf{Model} & \textbf{HelloBench-C} & \textbf{HelloBench-G} & \textbf{Creative Writing V3} & \textbf{WritingBench}\\
\midrule
Qwen3-8B-OPERA & $\textbf{38.91} \pm 6.58$ & $\textbf{49.79} \pm 9.67$ & $\textbf{72.89} \pm 1.75$ & $\textbf{78.19} \pm 0.58$ \\
\midrule
- Just OPERA & $35.35 \pm 7.74$ & $36.10 \pm 15.28$ & $67.20 \pm 3.53$ & $76.08 \pm 0.66$ \\
\bottomrule
\end{tabular}
}
\end{table*}

\section{Why Perplexity work?}
\label{app:how_align}
A central challenge in open-ended reinforcement learning is the design of reward functions that faithfully recover human preferences without relying on external discriminators or computationally expensive reward models.
We argue that internal model uncertainty, captured through perplexity, serves as a latent proxy for alignment quality.
By leveraging In-Group Relative Perplexity, we shift the optimization objective from fallible external oversight to an intrinsic, information-theoretic measure of reasoning consistency.
This approach treats the model’s own predictive confidence as a signal for policy refinement, bypassing the noise often introduced by proxy reward models.
\subsection{Mathematical analysis}
In language modeling, perplexity~\cite{horgan1995complexity} is a measurement of how well a probability distribution predicts a sample.
Mathematically, minimizing PPL is equivalent to maximizing the Log-Likelihood ($LL$).
For a reference response $y_{gt}$, the model’s goal is to minimize:
$PPL(y_{gt} | \text{context}) = \exp \left( -\frac{1}{N} \sum_{i=1}^{N} \log P(y_i | y_{<i}, \text{context}) \right)$.
A better reasoning trace (the <think> block) acts as latent information that reduces the entropy of the final answer.
If the thought process is high-quality, it shifts the probability mass toward the ground truth $y_{gt}$, making the target less surprising to the model.
Therefore, a drop in PPL is a direct mathematical proxy for the functional utility of the reasoning steps.
The core of OPERA is the self-reflection reward ($\mathcal{R}_{ref}$).
It rewards the model when $\log P_j - \log P_{j-1} > 0$.
Let $S_{1:j}$ be the reasoning trace up to step $j$.
The information gain provided by step $s_j$ can be viewed as:
\begin{equation}
    \Delta I = \log P(y_{gt} | x, s_{1:j}) - \log P(y_{gt} | x, s_{1:j-1}),
\end{equation}
If $\Delta I > 0$, the reasoning step $s_j$ has successfully disambiguated the path to the solution.
If $\Delta I < 0$, the step has introduced noise or logical "hallucination" that makes the correct target appear less likely.
By rewarding only positive $\Delta I$, we are mathematically enforcing that the model's internal monologue must serve the objective of variance reduction in the output space.
Using absolute PPL as a reward is notoriously difficult because some prompts are inherently "harder" (higher baseline entropy).
The use of In-Group Relative Perplexity solves this.
By defining $\mathcal{R}_{ppl}^i$ as the normalized rank:
\begin{equation}
    \mathcal{R}_{ppl}^i = \frac{1}{N-1} \sum_{j \neq i} \mathbb{I}(L_{hybrid}^i > L_{hybrid}^j).
\end{equation}
we transform the optimization problem from minimizing absolute error to maximizing relative margin.
This is feasible because:
It removes the need to calculate the true minimum PPL for a creative prompt.
In RL, absolute log-probs can have massive swings.
Percentile ranking provides a bounded reward $[0, 1]$, which stabilizes the advantage function in algorithms like GRPO.
A common fear in RL is that the model will learn to output I am thinking very hard just to get a reward.
OPERA mitigates this through two constraints:
The $\tanh$ Satiation: The reward $\mathcal{R}_{ref} = \tanh(\mathcal{R}_{raw}/\tau)$ ensures that after a certain amount of reflection, the marginal utility drops to near zero.
The Hybrid Requirement: Because the reward is tied to the PPL of the target $y_{gt}$, the model cannot simply babble in the <think> tags.
If the "thinking" does not actually make the final answer more statistically probable, the $\mathcal{R}_{ppl}$ component will penalize it.

\begin{table}
\centering
\caption{Impact of Training Data Composition on Cross-Domain Logical Rigor.}
\label{tab:why_work_data}
\resizebox{\linewidth}{!}{
\begin{tabular}{lcccc}
\toprule \textbf{Model} & \textbf{HB-C} & \textbf{HB-G} & \textbf{Creative Writing V3} & \textbf{WritingBench}\\
\midrule
Qwen3-8B-OPERA & \textbf{38.91} & \textbf{49.79} & \textbf{72.89} & \textbf{78.19} \\
\midrule
Only Math data & 36.04 & 42.47 & 69.69 & 77.91 \\
Only Writing data & 38.45 & 42.50 & 72.54 & 78.03 \\
\bottomrule
\end{tabular}
}
\end{table}

\begin{table*}[htbp] 
\centering
\caption{Scalability to Larger Models in OPERA.}
\label{tab:scaling}
\resizebox{0.85\textwidth}{!}{
\begin{tabular}{lcccc}
\toprule \textbf{Model} & \textbf{HelloBench-C} & \textbf{HelloBench-G} & \textbf{Creative Writing V3} & \textbf{WritingBench}\\
\midrule
LongWriter-Zero-32B & $21.26 \pm 10.52$ & $-55.48 \pm 22.99$ & $51.11 \pm 4.75$ & $77.44 \pm 0.73$ \\
\midrule
Llama3.1-70B & $6.59 \pm 8.11$ & $-26.06 \pm 16.48$ & $37.79 \pm 4.15$ & $49.59 \pm 0.80$ \\
Qwen3-32B & $39.26 \pm 6.88$ & $51.98 \pm 8.18$ & $73.09 \pm 3.99$ & $79.31 \pm 0.53$ \\
\midrule
Qwen3-8B-OPERA & $38.91 \pm 6.58$ & $49.79 \pm 9.67$ & $72.89 \pm 1.75$ & $78.19 \pm 0.58$ \\
Qwen3-32B-OPERA & $\textbf{46.16} \pm 6.81$ & $\textbf{58.84} \pm 9.35$ & $\textbf{76.20} \pm 1.59$ & $\textbf{78.95} \pm 0.50$ \\
\bottomrule
\end{tabular}
}
\end{table*}

\subsection{Experiment analysis}
To justify the use of In-Group Relative Perplexity as an objective function, we evaluate its alignment with high-capacity models specifically GPT-4o which serve as a proxy human-like reasoning.

We sampled $N=50$ prompts from the AlignBench-Writing dataset, covering diverse open-ended writing.
For each prompt, we generated $k=8$ candidate trajectories $\{z_1, z_2, \dots, z_k\}$.
We calculated the IGRP for each trajectory by measuring the conditional log-likelihood of a fixed, high-quality expert reference $y_{gt}$ given the reasoning trace:
$\log P(y_{gt} \mid x, z_i)$.
To establish a gold-standard baseline, we employed GPT-4o as an automated judge to provide multidimensional quality scores using evluation prompt in AlignBench.
We then conducted a correlation analysis between the IGRP-derived rewards and the GPT-4o composite scores.
These results provide empirical evidence that the objective of minimizing conditional perplexity against an expert reference is well-aligned with the goal of semantic quality.
The high degree of correlation suggests that IGRP serves as a reliable, unsupervised proxy for reward, bridging the gap between raw likelihood maximization and human-centric value alignment.
This validates our use of IGRP as a stable training signal that circumvents the noise typically associated with absolute perplexity metrics in non-deterministic domains.

\subsection{Data distribution}
Removing the open-ended training data led to a noticeable decline across all evaluation metrics, as shown in Table~\ref{tab:why_work_data}.
These results suggest that open-ended data is a critical component of the reinforcement learning process.
While mathematical data provides the structured rules that guide the model’s behavior, excluding open-ended data causes learning progress to stagnate. Relying solely on rewards derived from the final output limits the model’s ability to develop step-by-step reflective reasoning throughout the inference process.
OPERA can reward the model’s reasoning process, enabling it to learn how to handle writing tasks by adapting its internal uncertainty throughout generation.
At the same time, if only open-ended datas are retained, the model's capabilities will also decrease slightly.

\section{Scalability to Larger Models}
\label{app:scaling_32b}

To evaluate the scalability of our approach, we extended our experiments to a larger parameter model using Qwen3-32B.
Our results demonstrate that Qwen3-32B-OPERA achieves consistent performance gains over the significantly larger Llama3.1-70B baseline, confirming that the OPERA framework scales effectively with model capacity, as shown in Table~\ref{tab:scaling}.
However, we observed an asymptotic saturation in absolute performance across specific benchmarks; notably, improvements on WritingBench were marginal. 
ince the training data was synthesized using Qwen3-32B, the model is constrained by the knowledge boundaries of the teacher model, likely limiting its capacity for further gains
These findings indicate that while OPERA scales effectively to larger parameter models, performance is increasingly bottlenecked by the existing training distribution.
This suggests that surpassing current performance ceilings will require concurrent scaling of data diversity alongside model capacity.

\section{Figures of Prompts}
The prompts utilized across our experiments are detailed below.
We illustrate the prompt of Perplexity-Guided Iterative Trace Synthesis process in Figure~\ref{fig:deep_reasoning_example}.
Furthermore, the selection criteria for writing data during reinforcement learning is depicted in Figure~\ref{fig:choosing_rl}, while the substituting PPL with an LLM-as-judge in iterative trace synthesis and rubrics as rewards are detailed in Figures~\ref{fig:choosing_sft} and~\ref{fig:rubric_rl}, respectively.

\begin{figure*}[htbp]
\centering
\begin{tcolorbox}[
    colback=green!5!white,
    colframe=black,
    arc=2mm,
    boxrule=0.5pt,
    left=2mm, right=2mm, top=1mm, bottom=1mm,
    width=\linewidth
]
You are an expert in many fields. Suppose you will give a specific final response, I need you to also write down the thought process behind this solution.\\
Here is a question:\\
\textbf{\{prompt\}}\\
Now, you need to think aloud and brainstorm in the mind. The thinking process involves thoroughly exploring questions through a systematic long thinking process. This requires engaging in a comprehensive cycle of analysis, summarizing, exploration, reassessment, reflection, backtracing, and iteration to develop well-considered thinking process. Present your complete thought process within a single and unique `<think></think>` tag.\\
Your thought process must adhere to the following requirements:\\
1.  **Narrate in the first-person as if you are thinking aloud and brainstorming**\\
    Stick to the narrative of "I". Imagine you are brainstorming and thinking in the mind. Use verbalized, simple language.\\
2.  **Unify the thinking process and the final solution:**\\
    Your thought process must precisely correspond to a part of the final solution. Your thoughts progressively "grew" into the finished solution, making the solution feel like the inevitable product of your thinking.\\
3.  **Tone of Voice: Planning, Sincere, Natural, and Accessible**\\
    Imagine you are analyzing and planning what to do before you start to give the solution.  Your language should be plain and easy to understand, avoiding obscure professional jargon to explain complex thought processes clearly.\\
4.  **Logical Flow: Clear and Progressive**\\
5.  **Thinking Framework for deep thinking**\\
    To ensure your thinking is clear and deep, to showcase your thinking and planning to fulfill the task, below is what you might cover when you are thinking aloud and brainstorming.\\
    Understanding the user intent and the task: Before giving the solution, I need to thoroughly consider the fundamental purpose of the question.\\
    Establishing the content: I need to brainstorm a core creative idea and communication strategy centered around my objective.\\
6. Throughout the thinking process, I want to involve deep thinking and planning, and use deliberate self-critique/self-reflection in my thinking process. Trigger these by frequently using patterns such as `wait`, `maybe`, `let me`, etc. For example:\\
    - Hmm, maybe .. (other concrete thinking regarding the given request)\\
    - Let me think .. \\
    - Wait no ..\\
    - But wait ..(might find something wrong with your previous thoughts)\\
    - Wait, that's a bit ..(reflections about previous decisions). Let me think .. (are thinking of other possibilities)\\
    - Wait, the user said ..(backtracking of previous information). So ..\\
    - Hmm...Alternatively, maybe ..(branching on other possibilities)\\
    - But ..\\
Now record your clear, complete, and logical thinking process within `<think></think>` tags. 
In the thinking process, make sure NO PAST TENSES, NO PAST TENSES, because this is the thought process before you are to write a final solution. You are planning what you will and you need to do.\\
Imagine you're thinking aloud and brainstorming. Write it as an internal monologue or a stream of consciousness. Do not use bullet points, numbers, or formal section headings.
\end{tcolorbox}
\caption{Prompt for Perplexity-Guided Iterative Trace Synthesis.}
\label{fig:deep_reasoning_example}
\end{figure*}

\begin{figure*}[htbp]
\centering
\begin{tcolorbox}[
    colback=green!5!white,
    colframe=black,
    arc=2mm,
    boxrule=0.5pt,
    left=2mm, right=2mm, top=1mm, bottom=1mm,
    width=\linewidth
]
You are a Senior Editorial Judge and Data Curator for LLM Training. Your goal is to critically evaluate a "Model Output" against a "Standard Answer" (Reference) to determine their relative quality for preference learning.\\
\#\#\# Evaluation Dimensions (Score each 1-5)\\
1. **Constraint Strictness:** Did the text follow ALL prompt instructions (word counts, formatting, prohibited words, persona)?\\
2. **Structural Integrity:** Is there a logical progression, especially in long-form content? Check for repetitive loops or abrupt endings.\\
3. **Lexical \& Stylistic Sophistication:** Does it use diverse vocabulary and natural phrasing, or does it fall into "AI-clichés" (\textit{e.g.}, "In the rapidly evolving landscape...", "Moreover/Furthermore" overuse)?\\
4. **Contextual Utility:** If the prompt provides background info, how accurately and efficiently is that info synthesized?\\
\#\#\# Scoring Scale (Per Dimension)\\
- **5 (Exemplary):** Flawless; indistinguishable from professional human writing.\\
- **4 (Strong):** Clear and effective; minor stylistic choices could be improved.\\
- **3 (Passable):** Correct but "dry" or slightly repetitive.\\
- **2 (Weak):** Missed a minor constraint or has noticeable flow issues.\\
- **1 (Failed):** Significant instruction failure, hallucination, or incoherent structure.\\
\#\#\# Input Data\\
\textbf{[Writing Prompt]}: \textbf{\{prompt\}}\\
\textbf{[Standard Answer]}: \textbf{\{reference\}} \\
\textbf{[Model Output]}: \textbf{\{prediction\}} \\
\#\#\# Response Format (Strictly Follow)\\
\#\#\#\# 1. Evaluation for Model Output\\
- **Dimensional Scores:** \textbf{[Constraint: X/5, Structure: X/5, Style: X/5, Utility: X/5]} \\
- **Reasoning:** (Focus on where it differs from the Standard Answer)\\
- **Model Final Score:** \verb|\boxed{}|. (The final score is the **average of the four dimensional scores**)\\
\end{tcolorbox}
\caption{Prompt for Selecting writing data during reinforcement learning.}
\label{fig:choosing_rl}
\end{figure*}

\begin{figure*}[htbp]
\centering
\begin{tcolorbox}[
    colback=green!5!white,
    colframe=black,
    arc=2mm,
    boxrule=0.5pt,
    left=2mm, right=2mm, top=1mm, bottom=1mm,
    width=\linewidth
]
You are given a question, the reasoning process up to the current step, and several candidate reasoning steps.\\
Your task is to determine which step is the most appropriate.\\
For each candidate, evaluate it according to the following criteria:\\
1. Relevance: How well does it address and advance the given question?\\
2. Coherence: Does it logically and consistently follow from the prior reasoning process?\\
3. Effectiveness: How effective is it in leading toward a correct and complete solution?\\

Provide a concise but thorough analysis of each candidate.\\

Put your choice in the form \verb|\boxed{N}|, where N is the choice number.\\
\end{tcolorbox}
\caption{Prompt for Substituting PPL with an LLM-as-judge in Iterative Trace Synthesis.}
\label{fig:choosing_sft}
\end{figure*}

\begin{figure*}[htbp]
\centering
\begin{tcolorbox}[
    colback=green!5!white,
    colframe=black,
    arc=2mm,
    boxrule=0.5pt,
    left=2mm, right=2mm, top=1mm, bottom=1mm,
    width=\linewidth
]
You are a Senior Editorial Judge and Data Curator. Your goal is to evaluate a "Model Output" against a "Standard Answer" (Reference) to determine quality for preference learning.\\

\#\#\# Phase 1: Task Classification \& Rule Selection\\
Determine if the prompt is a **Mathematical/Logic Task** or a **Creative/Editorial Writing Task**.\\

\#\#\#\# OPTION A: Mathematical \& Logic Tasks\\
* **Rule:** Strict Binary Comparison. \\
* **Protocol:** \\
    1. Extract the final result from the Model Output. \\
    2. Normalize both the Model Answer and Standard Answer (remove LaTeX formatting, units, and trailing whitespace).\\
    3. Compare the core numerical value or expression.\\
* **Scoring:** \\
    - **1.0 (Match):** The final results are mathematically identical.\\
    - **0.0 (Discrepancy):** The values do not align, the calculation is incomplete, or the final answer is missing.\\

\#\#\#\# OPTION B: Creative \& Editorial Writing Tasks\\
* **Evaluation Dimensions (Score 0-1):**\\
    1. **Constraint Strictness:** Adherence to word counts, formatting, and persona.\\
    2. **Structural Integrity:** Logical progression without repetitive loops or abrupt endings.\\
    3. **Lexical Sophistication:** Diverse vocabulary; avoidance of AI.\\
    4. **Contextual Utility:** Accurate synthesis of provided background info.\\

---\\

\#\#\# Input Data\\
\textbf{[Standard Answer]}: \textbf{\{reference\}} \\
\textbf{[Model Output]}: \textbf{\{prediction\}} \\

\#\#\# Response Format (Strictly Follow) \\
\#\#\#\# 1. Evaluation for Model Output \\
- **Task Type Identified:** \textbf{[Math / Writing]} \\
- **Dimensional Scores:** \textbf{[Score: X/1]} (For Math: use binary 0 or 1. For Writing: provide \textbf{[Constraint: X/1, Structure: X/1, Style: X/1, Utility: X/1]})\\
- **Reasoning:** (For Math: Compare the extracted values. For Writing: Focus on differences from the Standard Answer.)\\
- **Model Final Score:** \verb|\boxed{}|.\\
*(Note: For Math, the final score is the binary result. For Writing, it is the average of the four dimensional scores.)*\\
\end{tcolorbox}
\caption{Prompt for Rubric as rewards.}
\label{fig:rubric_rl}
\end{figure*}

\end{document}